\documentclass[11pt, a4paper]{article}
\usepackage[utf8]{inputenc}
\usepackage{graphicx}
\usepackage{amsmath}
\usepackage{booktabs}
\usepackage{geometry}
\usepackage{natbib}
\usepackage{hyperref}
\usepackage{caption}
\usepackage{subcaption}
\usepackage{listings}
\usepackage{color}
\usepackage{authblk} 
\usepackage{booktabs}
\usepackage{tabularx}
\usepackage{listings}
\usepackage{xcolor}

\lstdefinelanguage{json}{
    basicstyle=\ttfamily,
    numbers=left,
    numberstyle=\tiny,
    stepnumber=1,
    numbersep=5pt,
    showstringspaces=false,
    breaklines=true,
    literate=
     *{0}{{{\color{blue}0}}}{1}
      {1}{{{\color{blue}1}}}{1}
      {2}{{{\color{blue}2}}}{1}
      {3}{{{\color{blue}3}}}{1}
      {4}{{{\color{blue}4}}}{1}
      {5}{{{\color{blue}5}}}{1}
      {6}{{{\color{blue}6}}}{1}
      {7}{{{\color{blue}7}}}{1}
      {8}{{{\color{blue}8}}}{1}
      {9}{{{\color{blue}9}}}{1}
}

\hypersetup{
    colorlinks=true,
    linkcolor=blue,
    filecolor=magenta,      
    urlcolor=cyan,
    citecolor=blue,
}

\definecolor{codegreen}{rgb}{0,0.6,0}
\definecolor{codegray}{rgb}{0.5,0.5,0.5}
\definecolor{codepurple}{rgb}{0.58,0,0.82}
\definecolor{backcolour}{rgb}{0.95,0.95,0.92}

\lstdefinestyle{mystyle}{
    backgroundcolor=\color{backcolour},   
    commentstyle=\color{codegreen},
    keywordstyle=\color{magenta},
    numberstyle=\tiny\color{codegray},
    stringstyle=\color{codepurple},
    basicstyle=\footnotesize\ttfamily,
    breakatwhitespace=false,         
    breaklines=true,                 
    captionpos=b,                    
    keepspaces=true,                 
    numbers=left,                    
    numbersep=5pt,                  
    showspaces=false,                
    showstringspaces=false,
    showtabs=false,                  
    tabsize=2
}
\title{CALM: A Framework for Continuous, Adaptive, and LLM-Mediated Anomaly Detection in Time-Series Streams}
\author[1]{Ashok Devireddy\thanks{The first author's contribution was performed during a summer internship at Google in 2025.}}
\author[1]{Shunping Huang}
\affil[1]{Dataflow ML Team \\ \texttt{ashokrd@berkeley.edu}, \texttt{shunping@google.com}}
\date{\today}

\begin{document}

\maketitle

\begin{abstract}
The detection of anomalies in non-stationary time-series streams is a critical but challenging task across numerous industrial and scientific domains. Traditional models, trained offline, suffer significant performance degradation when faced with concept drift, where the underlying statistical properties of the data change over time. This paper introduces CALM (Continuous, Adaptive, and LLM-Mediated), a novel, end-to-end framework for real-time anomaly detection designed to address this challenge. CALM is built on the Apache Beam distributed processing framework and leverages the TimesFm foundation model for forecasting-based anomaly detection. The framework's novelty lies in two core contributions. First, it implements a closed-loop, continuous fine-tuning mechanism that allows the anomaly detection model to adapt to evolving data patterns in near real-time. Second, it introduces an ``LLM-as-a-Judge'' component—a Large Language Model that provides semantic, context-aware judgments on detected anomalies to curate a high-quality training dataset, deciding whether an anomaly represents transient noise or a meaningful pattern shift. We evaluate CALM on the comprehensive TSB-UAD benchmark. Our results demonstrate that the continuously fine-tuned model improves the ROC AUC score in most datasets compared to the static, pre-trained base model, validating the efficacy of our adaptive, LLM-guided approach to maintaining high-performance anomaly detection in dynamic streaming environments.
\end{abstract}

\section{Introduction}

Time-series anomaly detection is a fundamental task with profound implications across a wide array of domains, including financial markets, industrial manufacturing, and IT systems monitoring. The timely identification of anomalous events—observations that deviate significantly from expected behavior—can prevent substantial economic losses, preempt system failures, and ensure operational reliability. As data generation becomes increasingly continuous and voluminous, the analysis of these time-series streams has become paramount.

A primary challenge in real-world streaming environments is the phenomenon of \textit{concept drift}, where the statistical properties of the data stream evolve over time. This non-stationarity invalidates the core assumption of many classical machine learning models, which are trained on static, historical datasets. A model that performs well at deployment time may see its accuracy degrade rapidly as new patterns emerge, a problem that necessitates a shift from static modeling to adaptive, online learning paradigms.

The recent advent of large-scale, pre-trained foundation models for time-series, such as TimesFm, represents a significant advancement. Trained on vast and diverse datasets comprising billions of data points, these models exhibit impressive zero-shot forecasting capabilities, providing a powerful baseline for various downstream tasks. However, their generalist nature, while a strength for broad applicability, can be a limitation in specialized, dynamic domains. Research suggests that while these models can be adapted for anomaly detection, they may not outperform specialized approaches without further tuning, particularly when faced with domain-specific concept drift  \citep{Fu2024FinancialFineTuning}.

This paper proposes that the solution lies not in choosing between a generalist foundation model and a specialized, adaptive one, but in architecting a system that achieves the best of both. While significant research has advanced scalable stream processing \citep{Akidau2015TheDataflow}, the power of time-series foundation models \citep{Liang2024FoundationModels}, and the reasoning capabilities of LLMs \citep{Zheng2023LLMasJudge}, these powerful paradigms have largely evolved in parallel. The critical gap, which this paper addresses, lies in their architectural synthesis. We argue that the next frontier in intelligent systems is not a better isolated algorithm, but a more integrated, self-correcting architecture. The true innovation is a novel architectural synthesis of three powerful and distinct technological paradigms: (1) scalable, stateful stream processing for robust data handling; (2) time-series foundation models for powerful, out-of-the-box forecasting; and (3) Large Language Models (LLMs) as sophisticated reasoning agents. By integrating these components into a cohesive, closed-loop system, we create a new class of intelligent infrastructure capable of autonomous adaptation.

To this end, we introduce CALM (Continuous, Adaptive, and LLM-Mediated), a holistic framework that leverages the zero-shot power of a foundation model as a starting point but integrates it into a fully automated, adaptive pipeline that specializes it over time. The CALM framework makes the following primary contributions:
\begin{enumerate}
    \item \textbf{A Scalable Streaming Architecture:} An end-to-end anomaly detection system architected on Apache Beam, specifically designed to handle high-throughput, out-of-order data streams. It employs stateful processing primitives to ensure correctness and robustness in a distributed environment \citep{Akidau2015TheDataflow, BeamStateful}.
    \item \textbf{The LLM-as-a-Judge Mechanism:} A novel component that functions as an online, semantic concept drift detector. Moving beyond the typical use of LLMs for offline evaluation \citep{LLMAsJudgePatronus}, our judge actively curates the fine-tuning dataset by providing reasoned judgments on anomalies, distinguishing transient noise from genuine pattern shifts to guide the model's adaptation \citep{Feki2023MLOpsFeedback}.
    \item \textbf{A Closed-Loop Continuous Fine-Tuning Pipeline:} A fully automated MLOps loop \citep{Feki2023MLOpsFeedback, MLOpsGoogleCloud} that collects data deemed valuable by the LLM Judge, batches it for training, triggers a fine-tuning job, and enables the dynamic, in-flight deployment of the improved model. This allows the system to autonomously adapt to concept drift without human intervention.
\end{enumerate}

We validate the CALM framework through a rigorous experimental evaluation on the TSB-UAD benchmark, demonstrating that our adaptive, LLM-guided approach yields significant performance improvements over a static, pre-trained model.
\section{Related Work}

\subsection{Taxonomy of Time-Series Anomaly Detection}
A time series is a sequence of data points indexed chronologically \citep{Salehi2024}. Such series can be \textit{univariate}, consisting of a single variable observed over time, or \textit{multivariate}, comprising multiple variables recorded concurrently \citep{Liang2024FoundationModels}. Anomalies within these series are broadly categorized into three types: \textit{point anomalies}, which are single data points deviating from the norm; \textit{contextual anomalies}, which are normal in a global sense but abnormal within their local context; and \textit{collective anomalies}, which are sequences of data points that, as a group, represent an anomalous pattern \citep{Chandola2009AnomalyDetectionSurvey, Hawkins1980Anomalies}.

The methodologies for detecting these anomalies are diverse, but a common taxonomy classifies them into four main categories: forecasting-based, reconstruction-based, representation-based, and hybrid methods \citep{Salehi2024, Blazquez-Garcia2021}. The CALM framework falls squarely within the \textbf{forecasting-based} paradigm. The fundamental principle of this approach is to train a model to predict future values of the time series based on its recent history. The deviation between the model's forecast and the actual observed value, often termed the prediction error or residual, serves as the anomaly score \citep{Laptev2015EGADS, Malhotra2015LSTMAD}. A data point is flagged as an anomaly if this error surpasses a predefined threshold \citep{Blazquez-Garcia2021, Hundman2018DetectingSpacecraft}. Our work refines this by using the model's quantile forecasts to establish a dynamic and robust statistical threshold.

\subsection{The Landscape of Time-Series Foundation Models}
Inspired by the transformative success of LLMs in Natural Language Processing (NLP), a recent paradigm shift has occurred in time-series analysis towards the use of large, pre-trained foundation models, often called Time-Series Foundation Models (TSFMs) \citep{Liang2024FoundationModels, Wen2024FM4TS}. These models are trained on massive, heterogeneous time-series corpora, enabling them to learn a rich variety of temporal patterns and exhibit impressive zero-shot performance on downstream tasks \citep{Garza2024TimeGPT, Das2024TimesFM}.

A key architectural innovation enabling TSFMs is \textbf{patching}, where the input time series is partitioned into contiguous ``patches'' that are treated as tokens \citep{Nie2023PatchTST}. This reduces the sequence length processed by the transformer, improving computational efficiency and allowing the model to learn local semantic meaning \citep{Garza2024Moirai}. Several prominent TSFMs have emerged, each with distinct architectural philosophies. \textbf{TimesFm}, the model used in our framework, is a decoder-only model trained to predict the next patch based on preceding ones, a design that naturally accommodates variable-length inputs and longer output horizons \citep{Das2024TimesFM, GoogleTimesFM}. Similarly, \textbf{Chronos} tokenizes time-series values via quantization into a fixed vocabulary, treating forecasting as a pure language modeling problem within a decoder-only framework \citep{Ansari2024Chronos}. In contrast, \textbf{Moirai} uses a masked encoder-based architecture, akin to BERT, and introduces ``any-variate attention'' by flattening multivariate series into a single sequence \citep{Garza2024Moirai, DataloopMoirai}. Other models like \textbf{PatchTST} focus on channel-independent patching for multivariate series \citep{Nie2023PatchTST}, while commercial offerings like \textbf{TimeGPT} leverage transformer architectures with conformal prediction to provide robust uncertainty intervals \citep{NixtlaTimeGPT, Olivares2023TimeGPT}. A brief comparison is provided in Table \ref{tab:tsfm_taxonomy}.

\begin{table}[ht] 
\centering 
\caption{Comparison of Time-Series Foundation Models}
\label{tab:tsfm_taxonomy}

\begin{tabularx}{\textwidth}{@{}ll>{\raggedright\arraybackslash}X>{\raggedright\arraybackslash}X@{}}
\toprule
\textbf{Model} & \textbf{Core Architecture} & \textbf{Key Feature(s)} & \textbf{Primary Use Case} \\ \midrule
TimesFm & Decoder-only Transformer & Patching; Long output patches & Zero-shot Forecasting \\
Chronos & Decoder-only Transformer & Value Quantization; Language Model & Probabilistic Forecasting \\
Moirai & Encoder-only Transformer & Any-variate Attention; Mixture Dist. & Multivariate Forecasting \\
PatchTST & Encoder-only Transformer & Channel-Independent Patching & Multivariate Forecasting \\
TimeGPT & Transformer & Conformal Prediction & Commercial Forecasting/AD \\ \bottomrule
\end{tabularx}
\end{table}

While these models demonstrate powerful zero-shot capabilities, their generalist nature can be a limitation in specialized domains with unique data characteristics and concept drift \citep{Alnegheimish2024SigLLM}. Research indicates that for challenging tasks like anomaly detection, fine-tuning or linear probing on task-specific data is often necessary to achieve state-of-the-art performance \citep{Goswami2024MOMENT}. CALM is built on this premise: it leverages the zero-shot power of a TSFM as a strong starting point but integrates a continuous fine-tuning mechanism to specialize it for a dynamic target environment.

\subsection{LLM-based Approaches to Anomaly Detection}
The application of LLMs to time-series anomaly detection is a burgeoning field with several distinct methodologies emerging. These can be broadly categorized as follows:
\begin{enumerate}
    \item \textbf{Direct Prompting:} These methods convert the numerical time-series into a textual representation and directly query an LLM to identify anomalous points. For example, the ``Prompter'' pipeline within the SigLLM framework serializes the time series and asks the LLM to output the indices of any anomalies \citep{Alnegheimish2024SigLLM}. While conceptually simple, this approach has been found to be less effective than forecasting-based methods \citep{Alnegheimish2024SigLLM_Consensus}.
    \item \textbf{Forecasting-based Detection:} This paradigm, which CALM employs, uses an LLM as a forecasting engine. The LLM predicts future values, and anomalies are flagged based on a significant residual error between the forecast and the actual observations. The ``Detector'' pipeline in SigLLM is a prime example of this approach \citep{Alnegheimish2024SigLLM}.
    \item \textbf{Representation and Distillation:} More architecturally integrated approaches use the LLM as a powerful feature extractor. \textbf{AnomalyLLM} \citep{Yu2024AnomalyLLM} proposes a knowledge distillation framework where a smaller student network is trained to mimic the feature representations of a larger, frozen LLM-based teacher network. Anomalies are then detected by a large discrepancy between the student and teacher outputs. Similarly, \textbf{TriP-LLM} \citep{Yu2025TriPLLM} uses a frozen, pre-trained LLM to process patch-wise tokens from three distinct branches (capturing local, selected, and global features) before a lightweight decoder reconstructs the input for anomaly scoring.
    \item \textbf{Multimodal Approaches:} A cutting-edge direction involves converting time-series into images and leveraging Vision-Language Models (VLMs). The \textbf{VisualTimeAnomaly} benchmark \citep{Luo2025VisualTimeAnomaly} demonstrates that VLMs can be robust anomaly detectors, particularly for range-based anomalies, and are surprisingly resilient to missing data, showcasing the potential of treating anomaly detection as a visual reasoning task.
\end{enumerate}
CALM utilizes the forecasting-based paradigm but innovates by embedding this capability within a fully automated, adaptive loop. Its primary novelty lies not in the forecasting method itself, but in the integration of an LLM-based semantic filter to guide the continuous adaptation of the forecasting model.

\subsection{Online Learning and Adaptation to Concept Drift}
A fundamental limitation of models trained offline is their inability to cope with \textbf{concept drift}, the phenomenon where the underlying data distribution of a stream changes over time \citep{Gama2014SurveyConceptDrift}. This drift renders static models obsolete and necessitates \textbf{online learning} or continuous adaptation strategies to maintain performance \citep{Lu2018LearningUnderConceptDrift, Feki2023MLOpsFeedback}. Traditional approaches to drift detection can be categorized as statistical or performance-based. Statistical methods use tests like the Kolmogorov-Smirnov (KS) test to detect changes in the input data distribution \citep{Ditzler2015Learning}. Performance-based methods monitor the model's output, such as its error rate, to detect degradation \citep{Gama2014SurveyConceptDrift}.

Modern frameworks often use a hybrid approach. For instance, \textbf{ADDAEIL} \citep{Zhang2025ADDAEIL} combines statistical distribution tests (KS and Mann-Whitney U tests) with an analysis of the internal structure of its ensemble model to detect drift and selectively retrain degraded components. However, these quantitative methods have inherent limitations. Statistical tests can be overly sensitive, triggering unnecessary retraining in response to harmless noise, while performance-based metrics may fail to detect significant data drift if it does not immediately degrade the chosen metric \citep{Sobolewski2023DataDrift}.

The CALM framework's continuous fine-tuning loop is a direct and practical implementation of an adaptive strategy. Crucially, it replaces purely statistical drift detection with a \textit{semantic} one. The LLM-as-a-Judge performs a qualitative assessment of whether a pattern change is meaningful, mimicking the cognitive process of a human analyst and addressing the brittleness of purely quantitative drift detectors. This aligns with established MLOps principles for Continuous Integration (CI), Continuous Delivery (CD), and Continuous Training (CT) of machine learning systems, which emphasize automation, monitoring, and iterative deployment to combat model decay \citep{Feki2023MLOpsFeedback, MLOpsGoogleCloud, DailyDoseMLOps}. CALM operationalizes these principles in a fully automated, closed-loop system driven by the data stream itself.

\subsection{Large Language Models as Reasoning Agents in ML Pipelines}
The role of LLMs is rapidly expanding from standalone applications to integral components within larger computational systems. A significant body of research has explored the use of LLMs for data annotation and labeling, automating tasks that are traditionally manual, costly, and time-consuming \citep{Yu2024AnomalyLLM, Tan2024LLMDataAnnotation}. LLMs can generate not only categorical labels but also rich auxiliary information like rationales and confidence scores \citep{He2023LargeLanguage}.

More recently, the \textbf{``LLM-as-a-Judge''} paradigm has emerged, primarily for the task of \textit{evaluating} the outputs of other AI models \citep{Zheng2023LLMasJudge, LLMAsJudgeBedrock}. These LLM judges have demonstrated a remarkable ability to provide nuanced, context-aware assessments that align closely with human preferences and are more informative than traditional metrics \citep{LLMAsJudgePatronus}. This is an active area of research, with frameworks like \textbf{MCTS-Judge} \citep{Wang2025MCTSJudge} exploring how to enhance judge reliability for complex, reasoning-intensive tasks by incorporating structured search algorithms like Monte Carlo Tree Search.

The CALM framework represents a significant evolution of this concept, shifting the role of the LLM from a passive evaluator to an active controller. Instead of using an LLM for offline evaluation or one-shot annotation, we deploy it as an \textbf{active, online, semantic filter} within the data processing pipeline itself. The LLM is not merely labeling data; it is executing a sophisticated reasoning task to guide the learning process of the downstream anomaly detection model. This moves beyond data annotation toward intelligent data curation for online adaptation.

This approach effectively automates the role of a human domain expert. In traditional human-in-the-loop systems, an analyst is often required to validate alerts, distinguish true anomalies from false positives, and ultimately decide if a model needs retraining in response to a pattern shift \citep{Feki2023MLOpsFeedback}. The LLM-as-a-Judge in CALM performs this exact cognitive function. By analyzing the data context before and after an outlier and deciding whether it signifies a ``transient, one-off event'' or a ``sustained shift,'' the LLM is making a determination about concept drift. It acts as a cognitive automaton, automating a high-level reasoning task that is a critical bottleneck in conventional adaptive systems. This represents a step toward more autonomous, self-correcting ML systems.
\section{The CALM System Architecture}

\subsection{High-Level Architectural Overview}
\begin{figure}[h!]
    \centering
    \includegraphics[width=\textwidth]{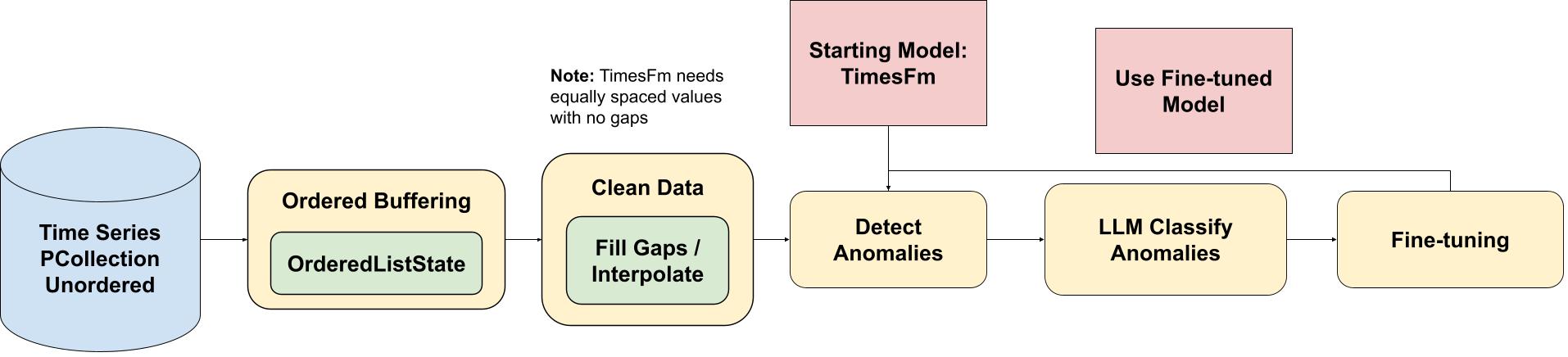}
    \caption{The high-level architecture of the CALM framework. Unordered data is processed through a series of stateful transformations, including ordered buffering and cleaning. Anomaly detection uses a dynamically swappable TimesFm model. The LLM Judge curates anomalies for the fine-tuning stage, which produces an improved model to be used in the next cycle.}
    \label{fig:high_level_overview}
\end{figure}

The CALM framework is an end-to-end streaming pipeline designed for scalability, robustness, and continuous adaptation. The high-level architecture, depicted in Figure \ref{fig:high_level_overview}, illustrates the closed-loop flow of data and models. The process begins with the ingestion of a potentially unordered stream of univariate time-series data points. This data undergoes stateful windowing and cleaning to prepare it for inference. Anomaly detection is performed using a dynamically loaded TimesFm model. Detected anomalies are then passed to the LLM-as-a-Judge for semantic classification. Anomalies deemed significant are collected and used to trigger a fine-tuning job, which produces an updated model. This new model is then seamlessly deployed back into the detection stage, completing the adaptive loop.

\subsection{Stateful Ingestion and Windowing on Apache Beam}
To handle the challenges of real-world data streams, which are often high-volume and delivered with out-of-order timestamps, the framework is built on Apache Beam, a unified model for defining both batch and streaming data-parallel processing pipelines.

\begin{figure}[h!]
    \centering
    \includegraphics[width=0.8\textwidth]{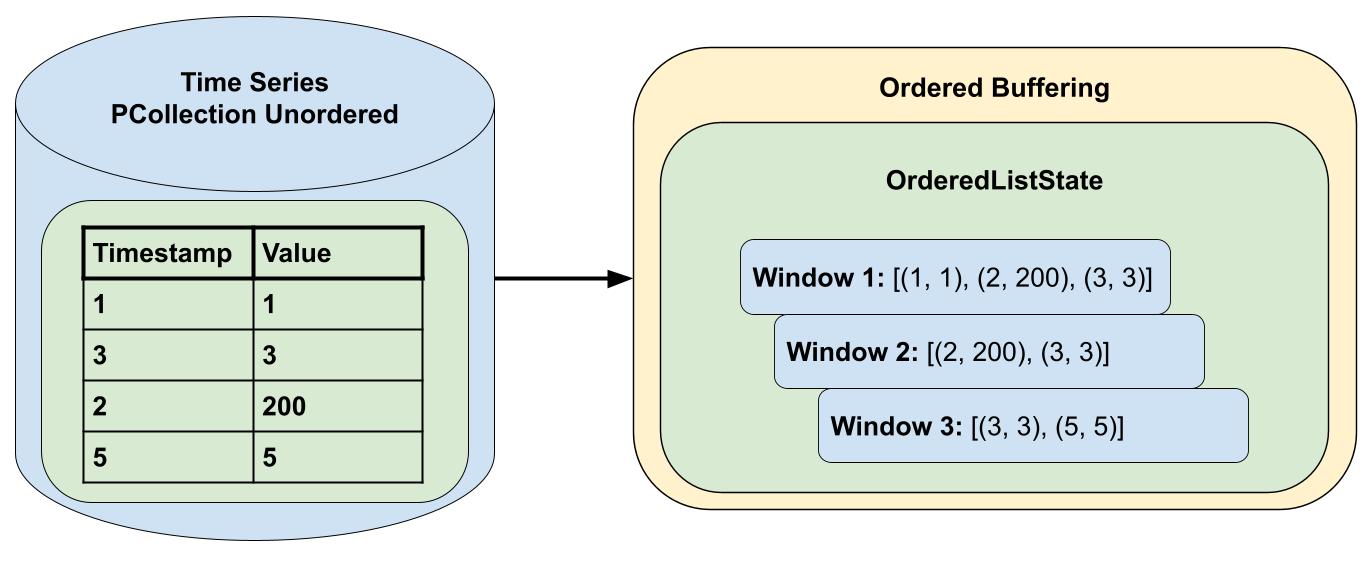}
    \caption{The process of ordered buffering. An unordered PCollection of time-series points is added to a stateful buffer (``OrderedListState''), which maintains the elements sorted by timestamp. This allows for the correct construction of overlapping sliding windows.}
    \label{fig:ordered_buffering}
\end{figure}

\paragraph{Handling Out-of-Order Data} 
A naive windowing approach would fail on out-of-order data, leading to incomplete or incorrect windows. We implement a custom stateful sliding-window function to handle this. As shown in Figure~\ref{fig:ordered_buffering}, incoming elements are first buffered in state (per key) instead of being immediately emitted. 

\begin{itemize}
  \item \textbf{Stateful Ordered Buffer:} We use a persistent in-memory buffer that maintains events sorted by their timestamp (leveraging Beam’s ordered state primitive). This ensures that even if elements arrive out of order (e.g., a timestamp 2 arrives after timestamp 3), they are placed in the correct order within the buffer.
  \item \textbf{Watermark Timer:} We employ a watermark-driven timer to determine when a window is ready to be emitted. The timer is set to fire at the end of each window interval, and it will only fire once the event-time watermark has passed the window’s end (indicating no more late data is expected for that interval). When the timer fires, the buffered events for that window are collected and emitted downstream as a complete window. After emission, the state buffer is cleaned up (removing old data), and a new timer is set for the next sliding window.
\end{itemize}

\begin{figure}[h!]
    \centering
    \includegraphics[width=\textwidth]{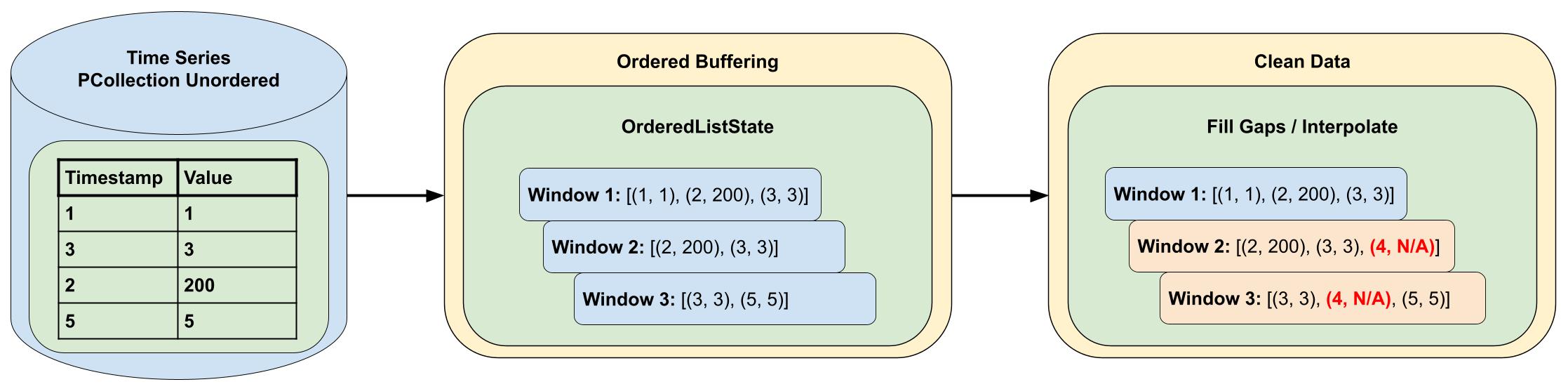}
    \caption{The data cleaning process. The gap-filling function takes a window of data and ensures it has a value for every expected timestamp. Missing values (e.g., at timestamp 4) are filled with a placeholder, preparing the data for the model, which requires regularly spaced inputs.}
    \label{fig:gap_filling}
\end{figure}

\paragraph{Data Cleaning and Interpolation} 
Foundation models like TimesFm need clean, regularly spaced data. We include a data-cleaning step that checks each window for missing timestamps based on the expected interval. As illustrated in Figure~\ref{fig:gap_filling}, this gap-filling function inserts a placeholder value (e.g., ``NaN") whenever a timestamp is missing, ensuring the window has a value at every expected time step. These placeholders can then be replaced using an appropriate interpolation strategy (e.g., linear or forward-fill) before the data is passed to the model.

\subsection{Dynamic Anomaly Detection with TimesFm}
\paragraph{Forecasting-based Detection} 
The core anomaly detection logic uses a forecasting model (TimesFm) on each window of data. We developed a custom inference component (integrated with Beam’s ``RunInference" transform) that processes each window by splitting it into a context portion and a prediction horizon. The model takes the context and produces a forecast for the horizon. Rather than using a single-point prediction, we leverage the model’s quantile forecasts (e.g., $q_{20}$, $q_{30}$, $q_{70}$, $q_{80}$) to establish a robust threshold for anomalies. We compute the interquartile range (IQR) from these quantiles, and flag a data point as anomalous if the actual value lies outside the range $[Q1 - 1.5 \times IQR,\; Q3 + 1.5 \times IQR]$. This statistical approach is less sensitive to non-Gaussian error distributions than one based on a fixed standard deviation.

\paragraph{Dynamic Model Swapping} 
A cornerstone of the CALM framework is its ability to adapt the model on the fly. We achieve this by using Apache Beam’s side input mechanism – an auxiliary, broadcast input that can provide configuration data to all workers. In our case, the side input continuously supplies the path of the latest fine-tuned model to the inference pipeline. The pipeline monitors a storage location (e.g., a local or remote bucket) for new model files. Whenever the fine-tuning stage produces a new model artifact and saves it to this location, the side input value is updated to the new model’s path. The inference component is designed to detect this update and then load the new model weights via a callback in the model handler. This entire process happens in-flight, allowing the streaming pipeline to seamlessly swap in the updated model without downtime.

\subsection{The LLM-as-a-Judge for Data Curation}
\begin{figure}[h!]
    \centering
    \includegraphics[width=\textwidth]{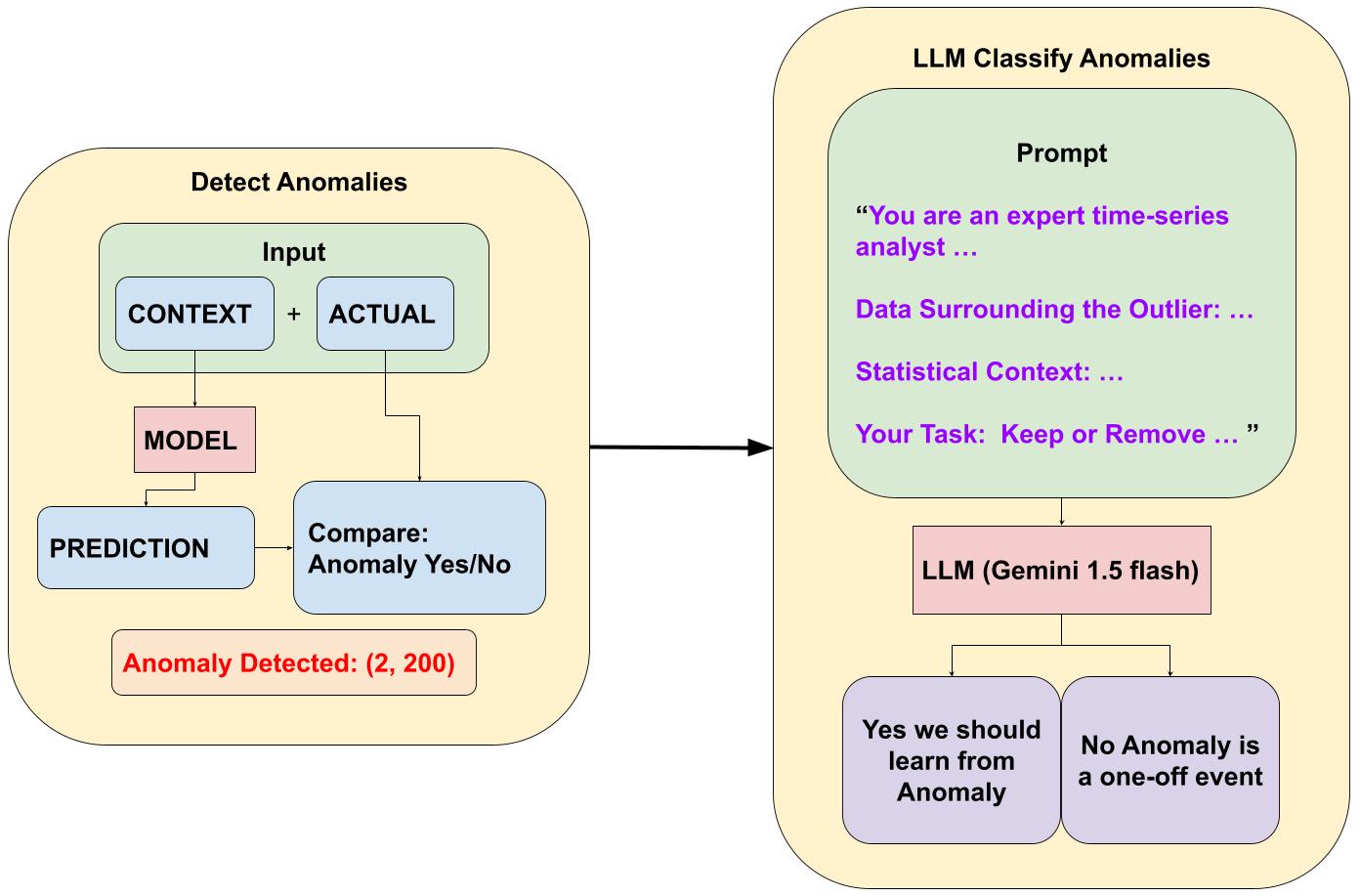}
    \caption{The LLM-as-a-Judge mechanism. A detected anomaly is enriched with surrounding data and statistical context. This is formatted into a detailed prompt and sent to an LLM (e.g., Gemini 1.5 Flash), which judges whether the anomaly represents a pattern to be learned (``KEEP'') or a one-off event to be ignored (``REMOVE'').}
    \label{fig:llm_judge}
\end{figure}

Once an outlier is detected by the forecasting model, it is not immediately used for fine-tuning. Instead, it is first passed to an LLM-based classifier (the "LLM-as-a-Judge" component) for semantic analysis, as depicted in Figure~\ref{fig:llm_judge}. 

\paragraph{Structured Prompting for Contextual Reasoning} The LLM-as-a-Judge constructs a detailed, structured prompt designed to provide the LLM with all necessary context to make an informed judgment. The prompt includes:
\begin{itemize}
    \item \textbf{Persona Setting:} ``You are an expert time-series analyst...'' to prime the model for the task.
    \item \textbf{Outlier Details:} The precise timestamp, actual value, predicted value, and anomaly bounds.
    \item \textbf{Surrounding Data:} A configurable number of data points immediately preceding and following the anomaly.
    \item \textbf{Statistical Context:} The mean and standard deviation of the data before and after the anomaly.
\end{itemize}
The LLM is then tasked to decide whether to ``KEEP'' the anomaly (if it signifies a sustained shift in the pattern) or ``REMOVE'' it (if it's a transient, one-off event).

\paragraph{Delayed Judgment with Stateful Context} 
A practical challenge arises if an anomaly occurs at the end of a window, leaving little “data after” for the LLM to consider. To handle this, our LLM-based judge uses a stateful approach to delay its decision until sufficient future context is available. In simpler terms, if an anomaly is detected but there are not enough subsequent points to provide context, the system will hold that anomaly aside temporarily. It buffers the anomaly and accumulates incoming data points from subsequent windows. A timer is set to defer the LLM evaluation for a short interval. Once enough new data has been collected to give a complete picture around the anomaly, the timer triggers and the LLM is invoked on the buffered data. This ensures the LLM always receives a full context (both before and after the outlier) before making its judgment.

\paragraph{Output and Routing} 
The LLM returns its verdict in a structured JSON format (including a rationale and a confidence score). If the decision is ``KEEP," the original data point (with its anomaly value) is forwarded to the fine-tuning pipeline. If the decision is ``REMOVE," we treat that point as a spurious outlier and replace its value with the model’s predicted value (effectively cleaning the anomaly from the data stream). In this way, only validated pattern shifts (labeled “KEEP”) are used to update the model, while isolated anomalies (“REMOVE”) are filtered out.

\subsection{Qualitative Analysis of the LLM Judge}
To provide insight into the reasoning capability of the LLM Judge, we present a qualitative example. Consider an anomaly detected in a dataset monitoring server CPU usage.

\textbf{Input to LLM:}
The anomaly is a sudden spike to 95\% CPU usage. The surrounding data shows usage typically hovering around 30\%, but in the period immediately following the spike, usage stabilizes at a new, higher baseline of 60\%.

\textbf{Generated Prompt Snippet:}
\begin{lstlisting}[language=bash]
...
**2. Data Surrounding the Outlier:**
* Data Before (25 points): [31.2, 29.8,..., 30.5]
* Data After (25 points): [62.1, 59.8, 61.5,..., 60.7]

**3. Statistical Context:**
* Mean Before: 30.15
* Mean After: 60.88
...
**4. Your Task:**
... Classify the outlier.
* REMOVE: If it is a transient, one-off event...
* KEEP: If it signifies a sustained shift in the pattern...
\end{lstlisting}

\textbf{LLM JSON Response:}
\begin{lstlisting}[language=json]
{
  "reasoning_steps": "The data after the outlier does not revert to the previous mean of ~30. Instead, it establishes a new, stable pattern around a mean of ~61. This indicates the outlier was not a transient error but the beginning of a new system state. This is a significant and sustained shift.",
  "decision": "KEEP",
  "confidence_score": 0.95
}
\end{lstlisting}
This example demonstrates the LLM's ability to go beyond the statistical fact of the outlier and perform contextual reasoning. It correctly identifies the sustained shift in the mean as evidence of a meaningful change, justifying its decision to ``KEEP'' the data point for fine-tuning. This qualitative result highlights the value of the LLM as a semantic filter.

\subsection{The Continuous Fine-Tuning Loop}
\paragraph{Batching for Fine-Tuning} 
The data points labeled “KEEP” by the LLM Judge are buffered and aggregated into batches for model training. We use a stateful batching mechanism that accumulates these points and only emits a batch when it has collected a full, continuous sequence of a predefined length. In other words, it waits until it can form a contiguous time-series segment (of size $N$, the batch size) before outputting it. This ensures that the data sent for fine-tuning constitutes a valid continuous window of time-series data, which is necessary for training the forecasting model.

\paragraph{Orchestrating the Fine-Tuning Job} 
When a batch of new data is ready, the system launches a fine-tuning process for the model. The first step is to determine the starting model: if a fine-tuned model from an earlier cycle exists, we use that; otherwise, we fall back to the original pre-trained model. Next, the batch of “KEEP” data is split into training and validation sets (for example, using an 80/20 split) to create a proper fine-tuning dataset. We then initiate a fine-tuning run on this data (using the TimesFm training routine). Upon completion, the updated model weights are saved as a new model artifact in persistent storage (for example, writing a file like `timesfm\_finetuned\_\{timestamp\}.pth' to a local or cloud storage bucket).

\paragraph{Closing the Loop} 
After the model is saved, the pipeline’s model-watching mechanism picks up the new model version. In our implementation, once the new model file appears in the storage location, it triggers an update to the side input discussed earlier. The anomaly detection stage then automatically loads the latest model for subsequent inferences. This closes the continuous learning loop: the system has now incorporated the newly fine-tuned model into production, and the pipeline returns to monitoring for the next anomalies and drift events.

\section{Experimental Evaluation}

\subsection{Benchmark and Dataset Selection}
To empirically validate the CALM framework, we conducted our evaluation on the \textbf{TSB-UAD benchmark} \citep{Paparrizos2022}, a comprehensive suite for unsupervised time-series anomaly detection. From the TSB-UAD v2 release, we selected a subset of \textbf{33 datasets} that met two specific criteria: (1) they contained multiple labeled anomaly groups, providing sufficient events for the model to potentially learn from, and (2) they had anomalous points present in the final 20\% of the series. This second criterion was essential to ensure we could fairly evaluate the fine-tuned model's performance on a strictly held-out set of future anomalies.

\subsection{Evaluation Methodology and Metric Justification}
Our experiment followed a strict temporal split to simulate a real-world scenario. For each dataset:
\begin{enumerate}
    \item The \textbf{first 80\%} of the data served as the ``live'' stream for fine-tuning.
    \item The \textbf{final 20\%} was a strict hold-out test set, used only for evaluation.
\end{enumerate}

The primary goal of CALM is not just to improve forecast accuracy but to enhance the model's fundamental ability to \textbf{distinguish between normal and anomalous states}. For this reason, we chose the \textbf{Receiver Operating Characteristic (ROC) Area Under the Curve (AUC)} score as our primary metric.

ROC AUC is ideal because it measures a model's discriminative power across all possible thresholds. This is crucial for validating our LLM-powered data curation. A key risk of naive fine-tuning is overfitting to noise; if a model learns to treat random, one-off spikes as normal, its forecast boundaries widen, and it fails to flag real anomalies. The LLM Judge is designed to prevent this by filtering out transient noise and retaining only anomalies that signify genuine pattern shifts. Therefore, a higher ROC AUC score is direct proof that the fine-tuned model has learned a more robust, generalizable understanding of ``normalcy,'' making it a superior anomaly detector.

\subsection{Results and Analysis}
The experiments demonstrate a clear and positive impact from the LLM-judged fine-tuning process.

\begin{figure}[h!]
    \centering
    \includegraphics[width=0.6\linewidth]{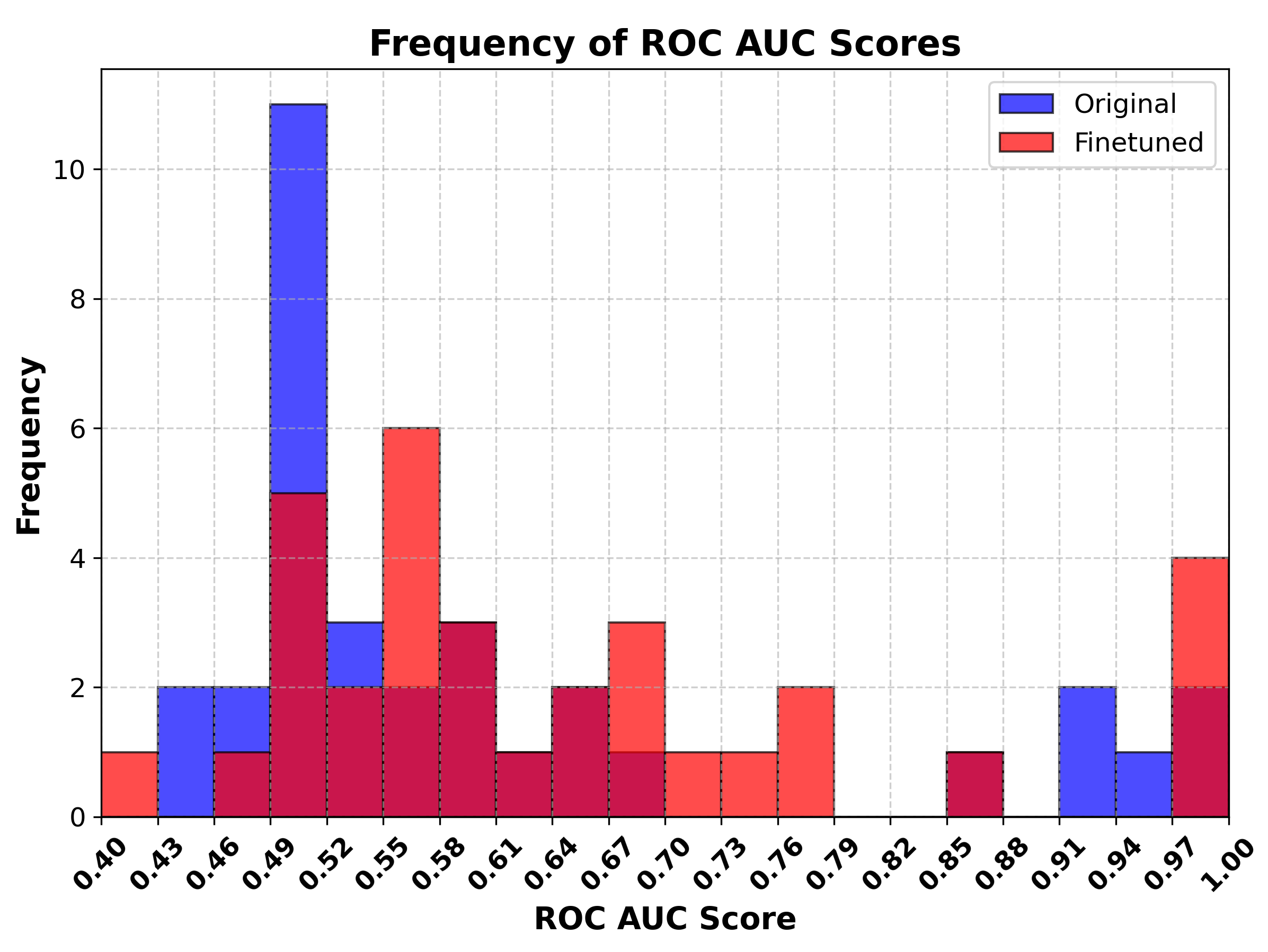}
    \caption{Distribution of ROC AUC scores across 45 TSB-UAD datasets. The panel shows overlaid histograms comparing the original and CALM-Tuned models. The visualization clearly illustrates that the LLM-guided fine-tuning process shifts the distribution of performance toward higher AUC values, with a more concentrated mass of datasets achieving scores between 0.6 and 1.0.}
    \label{fig:results_distribution}
\end{figure}

\begin{center} 
    \captionof{table}{Top and Bottom 10 Datasets by ROC AUC Improvement ($\Delta$)}
    \label{tab:combined_results}
    \begin{tabular}{lrrr @{\hspace{2em}} lrrr}
    \toprule
    \multicolumn{4}{c}{\textbf{Top 10}} & \multicolumn{4}{c}{\textbf{Bottom 10}} \\
    \cmidrule(r){1-4} \cmidrule(l){5-8} 
    \textbf{Dataset} & \textbf{Orig.} & \textbf{Tuned} & \textbf{Impr.} & \textbf{Dataset} & \textbf{Orig.} & \textbf{Tuned} & \textbf{Impr.} \\
    \midrule
    9.3..74\_col\_3    & 0.499 & 0.851 & +0.351 & WSD\_79            & 0.923 & 0.691 & -0.232 \\
    NAB\_data...\_4      & 0.503 & 0.748 & +0.245 & stock...0.25\_5     & 0.607 & 0.500 & -0.107 \\
    proc...-3-10\_col\_6 & 0.727 & 0.934 & +0.207 & stock...0.15\_2     & 0.879 & 0.788 & -0.091 \\
    8.3..73\_col\_2    & 0.568 & 0.774 & +0.207 & SWaT...\_col\_58    & 0.512 & 0.430 & -0.082 \\
    101-freeway-traffic & 0.454 & 0.648 & +0.194 & KPI-e0747cad...    & 0.480 & 0.423 & -0.057 \\
    6.1..65\_col\_5    & 0.439 & 0.600 & +0.161 & 9.3..74\_col\_1    & 0.592 & 0.552 & -0.040 \\
    WSD\_37              & 0.565 & 0.674 & +0.110 & stb-6              & 0.500 & 0.463 & -0.037 \\
    proc...-2-7\_col\_34 & 0.469 & 0.557 & +0.088 & WSD\_183            & 0.587 & 0.558 & -0.028 \\
    stb-5               & 0.504 & 0.583 & +0.079 & SWaT...\_col\_60    & 0.501 & 0.480 & -0.021 \\
    stb-32              & 0.456 & 0.527 & +0.071 & SED\_20000\_0       & 0.457 & 0.437 & -0.019 \\
    \bottomrule
    \end{tabular}
\end{center}

Table \ref{tab:combined_results} Top 10 shows that the framework can yield substantial gains, especially on datasets where the original model performed poorly (AUC near 0.5). This highlights its ability to adapt a general model to specific, previously challenging patterns. Conversely, Table \ref{tab:combined_results} Bottom 10 provides a balanced view, showing cases where performance decreased. This often occurred on datasets where the original model was already highly performant, suggesting the fine-tuning process may have slightly overfit to the limited anomalies in the training portion. This represents a clear area for future work, such as developing more intelligent triggers for the fine-tuning process or introducing noise before finetuning to prevent overfitting.

\subsection{Hyperparameter Sensitivity Analysis}
\begin{figure}[h!]
    \centering
    \begin{subfigure}[b]{0.48\textwidth}
        \centering
        \includegraphics[width=\textwidth]{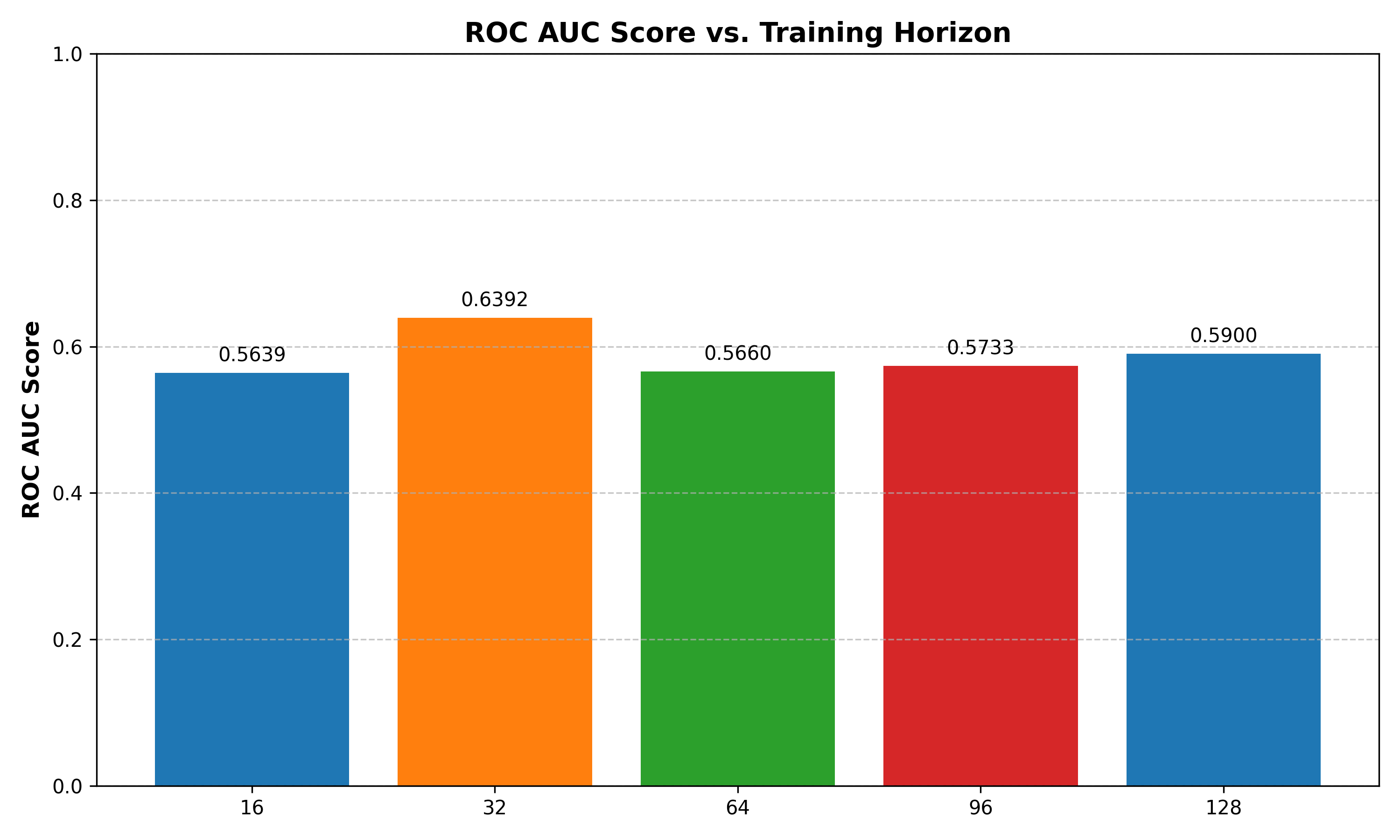}
        \caption{Impact of Prediction Horizon}
        \label{fig:horizon_roc_sub}
    \end{subfigure}
    \hfill 
    \begin{subfigure}[b]{0.48\textwidth}
        \centering
        \includegraphics[width=\textwidth]{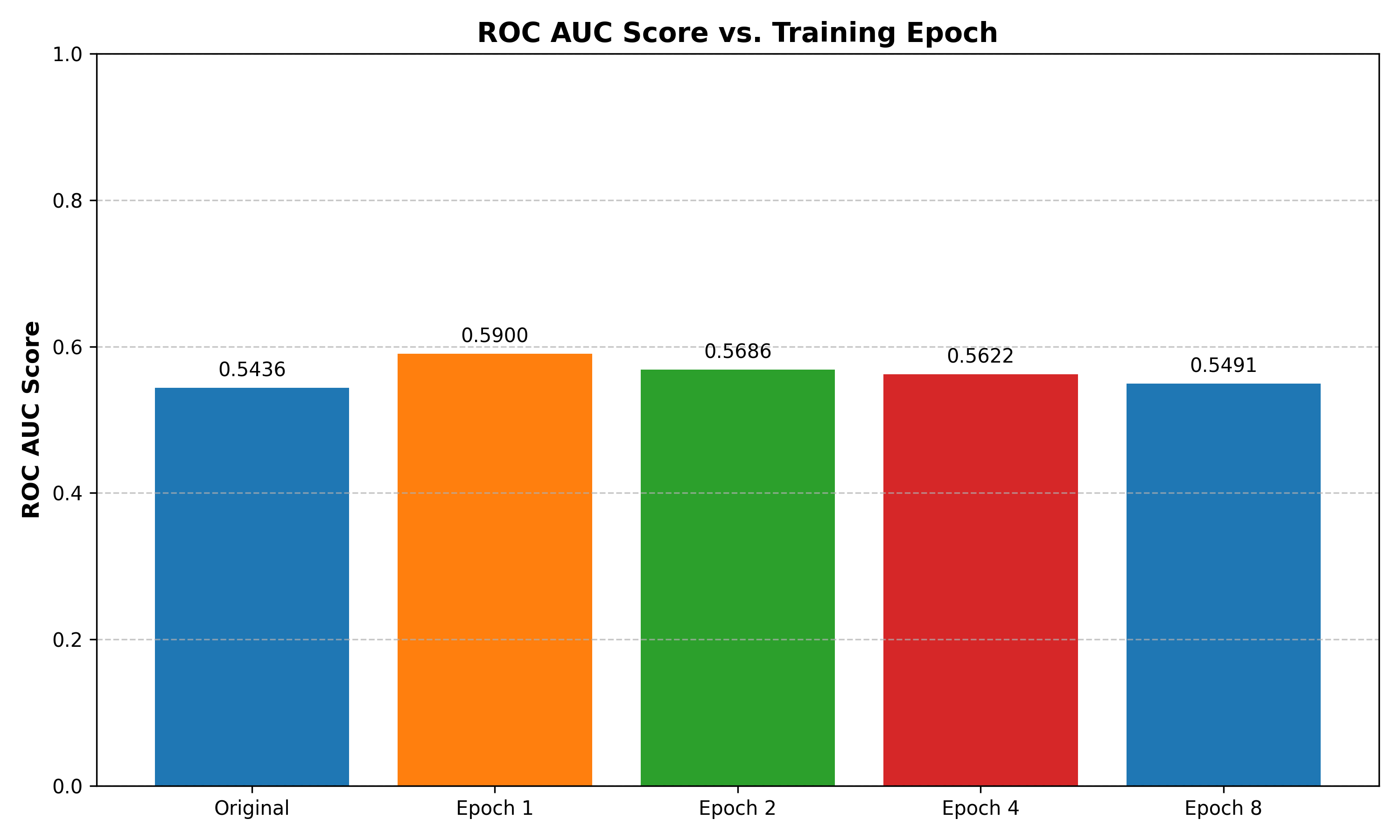}
        \caption{Impact of Training Epochs}
        \label{fig:epoch_roc_sub}
    \end{subfigure}
    \caption{ROC AUC score sensitivity to prediction horizon (a) and number of fine-tuning epochs (b) on the WSD\_94 dataset.}
    \label{fig:hyperparam_comparison}
\end{figure}

To better understand the behavior of the CALM framework, we conducted a sensitivity analysis on two key hyperparameters using the WSD\_94 dataset: the prediction horizon length and the number of fine-tuning epochs. These experiments, shown in Figure~\ref{fig:hyperparam_comparison}, provide insight into the framework's robustness and help identify optimal settings for deployment.

\paragraph{Impact of Prediction Horizon}
We first investigated how the model's prediction horizon affects its anomaly detection performance (Figure~\ref{fig:horizon_roc_sub}). We varied the horizon length, testing values of 16, 32, 64, 96, and the default of 128. The results indicate that the framework's performance is surprisingly robust to changes in this parameter. While a horizon of 32 yields the highest ROC AUC score of 0.6392, the scores for other horizons remain competitive. This phenomenon can likely be attributed to the TimesFM architecture, which is pre-trained to output a fixed-length forecast (e.g., 128 steps) that is subsequently truncated if asked for a shorter horizon.

\paragraph{Impact of Training Epochs}
Next, we examined the effect of the number of fine-tuning epochs (Figure~\ref{fig:epoch_roc_sub}). The most substantial performance gain is achieved within the very first epoch, where the ROC AUC score jumps from 0.5436 (original model) to 0.5900. This demonstrates the rapid effectiveness of the CALM framework's adaptation mechanism. However, subsequent epochs yield marginal improvements, suggesting that the model begins to overfit. This pattern indicates that a short fine-tuning cycle (e.g., one epoch) is both sufficient and optimal for adapting to concept drift without significant overfitting.

\section{Discussion and Future Work}
The CALM framework demonstrates a powerful architectural pattern for building next-generation, self-adapting monitoring systems. The fusion of scalable stream processing, pre-trained foundation models, and LLM-based reasoning agents provides a robust blueprint that could be extended beyond anomaly detection to other adaptive tasks like dynamic forecasting and online classification.

\paragraph{Limitations} Despite its promising results, the framework has several limitations that warrant discussion. First, the inclusion of LLM API calls and periodic fine-tuning jobs introduces both monetary \textbf{cost and processing latency}. There is an inherent trade-off between the frequency of adaptation (and thus responsiveness to drift) and the operational cost of the system. Second, the framework's performance is partially dependent on the \textbf{reliability and consistency of the LLM judge}. While powerful, LLMs can be sensitive to prompt phrasing and may occasionally produce inconsistent outputs, which could impact the quality of the fine-tuning data. This contrasts with the formal, albeit potentially brittle, statistical guarantees of methods like ADDAEIL \citep{Zhang2025ADDAEIL}. Third, the current implementation is focused on \textbf{univariate time-series}, a limitation inherited from the base TimesFm 1.0 model \citep{Das2024TimesFM}. This is a significant constraint, as many real-world systems require detecting anomalies based on inter-variable correlations \citep{Zhao2020Multivariate, Liang2024FoundationModels}.

\paragraph{Future Work} These limitations point to several exciting avenues for future research:
\begin{itemize}
    \item \textbf{Parameter-Efficient Fine-Tuning (PEFT):} To mitigate the cost and time of full fine-tuning, future versions could explore PEFT methods \citep{Lialin2023ScalingDown, PEFT_Kanerika}. By training only a small subset of the model's parameters, PEFT could enable more frequent and cheaper adaptation cycles. A particularly promising technique is \textbf{Low-Rank Adaptation (LoRA)} \citep{Hu2021LoRA}, which injects small, trainable low-rank matrices into the model. Recent work has already demonstrated the successful application of LoRA to time-series foundation models, making it a viable path for optimizing the CALM framework \citep{Bansal2024LoRA_TSFM}.
    \item \textbf{Extension to Multivariate Time-Series:} A significant extension would be to adapt the framework for multivariate anomaly detection. This would require replacing the univariate TimesFm with a multivariate-capable foundation model. Promising candidates include \textbf{Moirai}, which is designed for ``any-variate'' time series by flattening variables into a single sequence \citep{Garza2024Moirai}, or adopting architectures like \textbf{DualLMAD}, which uses parallel GPT-2 backbones to explicitly model temporal and inter-metric dependencies \citep{Zhang2024DualLMAD}. This extension would also necessitate developing more complex logic for the LLM Judge to reason about inter-variable relationships.
    \item \textbf{Advanced LLM Judging Strategies:} The judging component offers significant room for advancement. This includes upgrading the reasoning engine itself. While our work uses Gemini 1.5 Flash for efficiency, employing more powerful models could enhance reasoning capabilities. More importantly, the interaction logic could be improved. Inspired by frameworks like \textbf{MCTS-Judge} \citep{Wang2025MCTSJudge}, which uses Monte Carlo Tree Search to guide an LLM to more reliable conclusions in complex domains like code evaluation, a future version of CALM could implement a multi-step verification or chain-of-thought reasoning process for the judge. Furthermore, using an ensemble of different LLM judges could improve robustness and reduce the impact of idiosyncratic errors from a single model.
    \item \textbf{Optimizing the Fine-Tuning Trigger:} The current trigger for fine-tuning is based on collecting a fixed-size batch of data. A more intelligent trigger could be developed, for example, based on the rate of ``KEEP'' decisions from the LLM. A high rate would signal that significant concept drift is occurring, justifying an immediate fine-tuning cycle, thereby making the system's adaptivity more responsive to the dynamics of the data stream.
    \item \textbf{Robustness through Noise Augmentation:}
A critical issue observed in fine-tuning is that the model can overfit to new ``normal'' data, causing the predicted quantiles [$q_{0.1}$, $q_{0.9}$] to shrink. To investigate a solution, we conducted a preliminary experiment with \textbf{noise augmentation}, a regularization technique also known as jittering \citep{Wen2021TS_Augmentation}.

The \texttt{noise\_level} is a critical hyperparameter that controls the intensity of the Gaussian noise, $N(0, \sigma^2_{\text{noise}})$, added to the training data. To ensure the noise is appropriately scaled to the data's magnitude and variance, we define the standard deviation of the noise ($\sigma_{\text{noise}}$) as a fraction of the standard deviation of the entire training time series ($\sigma_{\text{data}}$). This relationship is formalized by the equation:
$$
\sigma_{\text{noise}} = \sigma_{\text{data}} \times \texttt{noise\_level}
$$
Therefore, for our experiment with a \texttt{noise\_level} of 0.01, the standard deviation of the injected noise is precisely 1\% of the standard deviation of the training data. This proportional scaling makes the augmentation strategy robust, as it automatically adapts the noise intensity to the specific statistical characteristics of any given time series.
We implemented a \texttt{NoisyTimeSeriesDataset} class that injects Gaussian noise into the training data on-the-fly. For each sample requested by the dataloader, noise is generated and added to the numpy arrays of the context and horizon windows. The standard deviation of this noise is proportional to the standard deviation of the entire training series, controlled by a \texttt{noise\_level} hyperparameter, which we set to 0.01 for this experiment. This on-the-fly approach ensures that noise is applied only to the training set and that the model sees a different random perturbation in each epoch, which is an effective regularization strategy \citep{Bishop1995NoiseRegularization}.

The results of this initial experiment are shown in Figure~\ref{fig:noise_epoch_comparison}. Without noise (the solid line), the model's performance peaks at Epoch 1 and then degrades, a classic sign of overfitting. When noise is introduced (the dashed line), the performance curve becomes flatter. While the peak performance is slightly lower, the degradation in later epochs is mitigated. This suggests that noise augmentation is successfully acting as a regularizer, forcing the model to learn more general features instead of memorizing the training data.

These findings are highly promising. A clear direction for future work is to perform a comprehensive hyperparameter search to identify the optimal \texttt{noise\_level}. A well-tuned noise level could not only prevent overfitting but also lead to a higher overall performance by allowing the model to train for more epochs without performance degradation.

\begin{figure}[h!]
    \centering
    \includegraphics[width=0.8\textwidth]{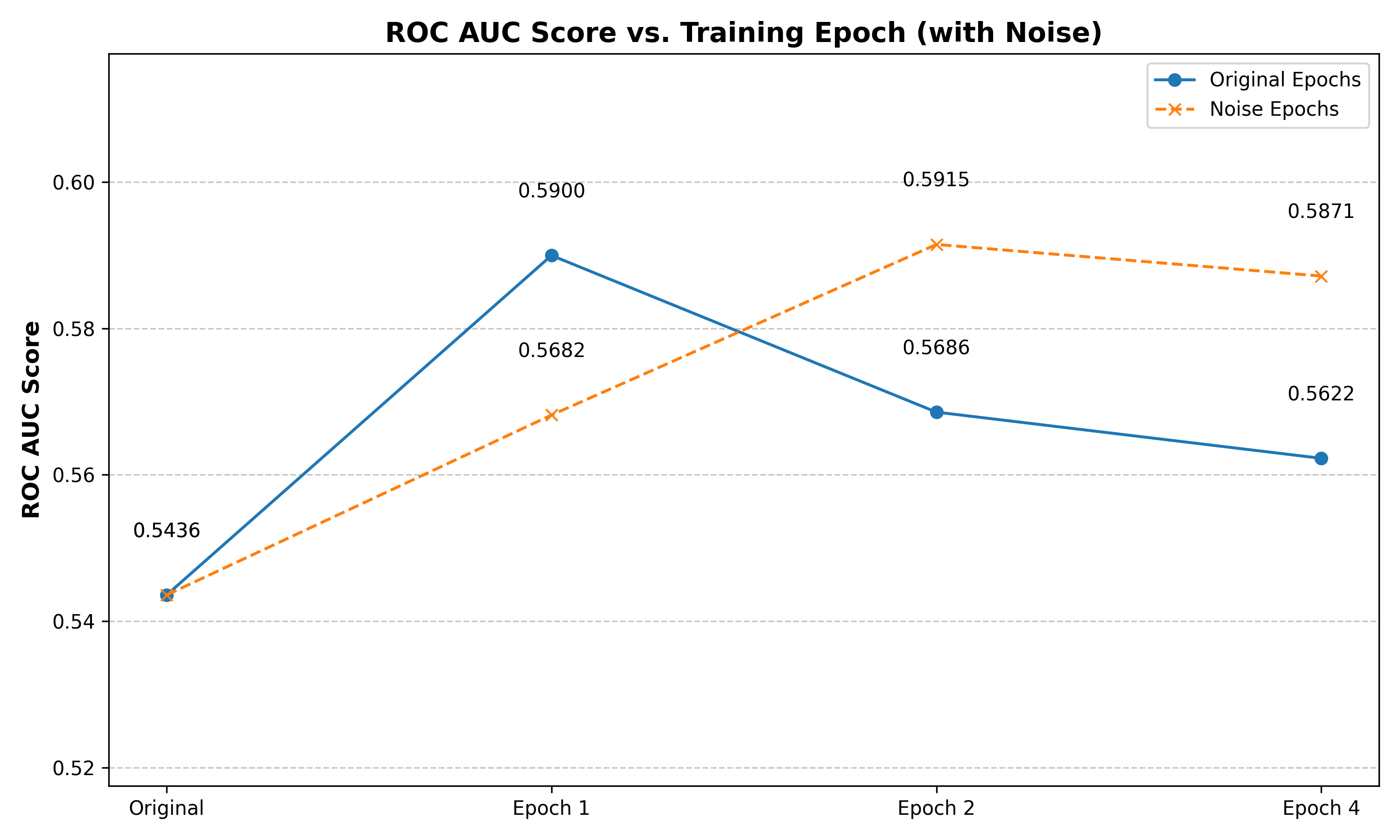}
    \caption{Comparison of ROC AUC scores over training epochs with and without noise augmentation (\texttt{noise\_level=0.01}). Noise acts as a regularizer, mitigating the performance degradation from overfitting seen in later epochs.}
    \label{fig:noise_epoch_comparison}
\end{figure}
\end{itemize}

\section{Conclusion}
This paper addressed the critical challenge of anomaly detection in non-stationary time-series streams. We introduced CALM, a novel framework that provides a robust and scalable solution by synthesizing three key technologies. It leverages Apache Beam for stateful stream processing, the TimesFm foundation model for powerful forecasting, and a Large Language Model as an intelligent, semantic judge to curate data for adaptation. The core of the framework is a fully automated, closed-loop continuous fine-tuning pipeline that allows the system to autonomously adapt to concept drift. Our empirical evaluation on the TSB-UAD benchmark confirmed that this adaptive, LLM-guided approach leads to significant performance gains over a static, pre-trained model. The CALM architecture serves as a blueprint for a new class of intelligent, self-adapting monitoring systems capable of operating effectively in dynamic, real-world environments.

\section*{Acknowledgments}

The authors would like to express their gratitude to the Dataflow ML team at Google for their invaluable support and collaboration. 
Special thanks go to Xiangqian Hu for guidance, Claude van der Merwe for co-hosting and mentorship, and Danny McCormick for his insightful feedback. 
The first author also wishes to thank his co-author, Shunping Huang, for serving as his host during the internship and for his continual support throughout the project.

\bibliographystyle{plainnat}
\bibliography{references}

\end{document}